\documentclass[journal]{IEEEtran}

\usepackage{graphicx}
\usepackage{comment}
\usepackage{amsmath,amssymb}
\usepackage{color}
\usepackage{subcaption}
\usepackage{booktabs}
\usepackage{multirow}
\usepackage{array}
\usepackage{cite}

\definecolor{ground}{RGB}{127, 201, 127}
\definecolor{pred}{RGB}{190, 174, 212}

\usepackage{fixltx2e}
\usepackage{stfloats}

\ifCLASSOPTIONcaptionsoff
  \usepackage[nomarkers]{endfloat}
 \let\MYoriglatexcaption\caption
 \renewcommand{\caption}[2][\relax]{\MYoriglatexcaption[#2]{#2}}
\fi

\usepackage{moreverb,url}
\usepackage[colorlinks,bookmarksopen,bookmarksnumbered,citecolor=red,urlcolor=red]{hyperref}

\usepackage{xcolor}

\definecolor{megenta}{RGB}{244, 35, 234}
\definecolor{darkpurple}{RGB}{128, 64, 128}
\definecolor{lightgreen}{RGB}{152, 251, 152}

\begin{document}

\title{OASIS: Automated Assessment of Urban Pedestrian Paths at Scale}

\author{Yuxiang~Zhang,
        Suresh~Devalapalli,
        Sachin~Mehta,
        and~Anat~Caspi 
\thanks{Preprint. Under review.}
\thanks{Corresponding author: Yuxiang Zhang (email: yz325@uw.edu)}
}

\maketitle

\begin{abstract}
The inspection of the Public Right of Way (PROW) for accessibility barriers is necessary for monitoring and maintaining the built environment for communities' walkability, rollability, safety, active transportation, and sustainability. However, an inspection of the PROW, by surveyors or crowds, is laborious, inconsistent, costly, and unscalable. The core of smart city developments involves the application of information technologies toward municipal assets assessment and management. Sidewalks, in comparison to automobile roads, have not been regularly integrated into information systems to optimize or inform civic services. We develop an Open Automated Sidewalks Inspection System (OASIS), a free and open-source automated mapping system, to extract sidewalk network data using mobile physical devices. OASIS leverages advances in neural networks, image sensing, location-based methods, and compact hardware to perform sidewalk segmentation and mapping along with the identification of barriers to generate a GIS pedestrian transportation layer that is available for routing as well as analytic and operational reports. We describe a prototype system trained and tested with imagery collected in real-world settings, alongside human surveyors who are part of the local transit pathway review team. Pilots show promising precision and recall for path mapping (0.94, 0.98 respectively). Moreover, surveyor teams'  functional efficiency increased in the field. By design, OASIS takes adoption aspects into consideration to ensure the system could be easily integrated with governmental pathway review teams' workflows, and that the outcome data would be interoperable with public data commons.
\end{abstract}

\begin{IEEEkeywords}
Pedestrian mapping, First-last-mile, Image processing, Object detection, Semantic segmentation
\end{IEEEkeywords}


\IEEEPARstart{S}{idewalks} are at the heart of the public built environment, offering a mode of travel that connects nearly all other travel options and community activities. However, sidewalks are not safe and accessible to all. Many travelers, particularly those with disabilities, encounter sidewalk obstacles, like steep inclines, damaged concrete, or curb barriers, that limit their access to the benefits of urban life. Detailed networked information about the paths, connectivity of paths, avoidable obstacles, and other amenities along paths, provide pedestrians with different mobility concerns the ability to discover routes appropriate for them, plan trips and foreshadow what they may find on the ground  \cite{quinones2011supporting, zimmermann2021independent, bosch2017flying}. Additionally, in the United States, under section 504 of the Rehabilitation Act (1973) and Title II of the Americans with Disabilities Act (ADA) (1990) \cite{ADA}, any public entity subject to ADA federal regulation must perform self-evaluation of their public right of way (PROW) to identify barriers in the paths and specify a remediation plan.
While driver-routing, asset management, and transportation planning have motivated extensive mapping, surveying, and monitoring of automobile roads (e.g., \cite{cordts2016cityscapes, varma2019idd, sun2020scalability, mi2021hdmapgen}), pedestrian street-side environments have not been regularly surveyed or monitored\cite{bolten2017pedestrian}. Studies in the U.S. focusing on the PROW, which consists of civic environments and paths for public use (including sidewalks and footpaths), have demonstrated that government agencies often lack knowledge about their own PROW and the barriers within those pathways\cite{eisenberg2020communities, eisenberg2022barrier}. Importantly, both public and private stakeholders need and would benefit from routable graphs describing sidewalk networks. Public stakeholders require data collection and monitoring of sidewalk environments including first responders in routing to emergent settings, road construction coordinators, Safe Route to Schools programs, Vision Zero programs, as well as paratransit teams to manage door-to-door transportation and determine rider eligibility. Private uses involve shipping and freight applications identifying curbside-to-door routes, and street-side autonomous navigating devices, e.g. delivery robots and autonomous wheelchairs. There is evidence to support the idea that the information gap between the cartographic information needed for pedestrians with disabilities to navigate in the sidewalks and the existing mapped information severely deteriorates the performance of transportation networks and impacts the security and safety of travelers \cite{gallo2010augmented, pyun2013advanced, USGAO2007}. Our goal is to introduce a system for automated and ongoing monitoring of pedestrian path networks and an analytic software solution that can leverage streetside footage to obtain useful, open, shared pedestrian network data that is geographically accurate, consistent, and usable by both public and private stakeholders. The desired outcome is open, shared, interpretable data that enhances available tools, visualizations, and metrics to manage transportation networks, accessibility, and traversability at scale \cite{bolten2021towards}. 

We present an open-source system and proof of concept implementation tested through a partnership with King County Metro Access, which serves as an applied case study environment. Our pilot was tested in three different municipal environments to demonstrate the feasibility of system adoption in real-world environments, and in the hands of real-world agencies.

The system contribution, Open Automated Sidewalks Inspection System (OASIS), is an analytic system that provides inferred sidewalk network output via a human-guided device composed of direct-to-consumer off-the-shelf electronics. It consists of a number of steps: OASIS captures street-view images, segments infrastructure and street furniture in the scene, and generates mapping information for sidewalk assessment in real-time-- potentially avoiding image data storage, per user preference. OASIS is built on a low-power and lightweight edge device to enable easy integration into other powered mobility devices. The pedestrian paths assessment process which is typically done by human surveyors is automated and confers an efficient and scalable way to assist with sidewalk mapping and review. With OASIS, we demonstrate the ability to generate pathway assessment data for multiple municipalities, in a standardized format that can be reused by other stakeholders or consumed by downstream applications.

We introduce the work and case study in the following sections. Section \ref{sec:case_study_partner} details the real-world setting and organizational partner for our pilot. Section \ref{sec:related_work} discusses related and prior work in outdoor transportation environment mapping. Section \ref{sec:mapping_system} details the implementation and methods utilized in building our system. Section \ref{sec:hardware} describes the hardware used in the preliminary proof of concept implementation. Quantitative and qualitative results from the pilot study are provided in Section \ref{sec:results}, followed by discussion and conclusions in Section \ref{sec:conclusion}.

\section{Case study stakeholders}
\label{sec:case_study_partner}
Transportation agencies are federally mandated to provide ADA-complementary paratransit service. These are transportation services provided to individuals with disabilities to supplement fixed-route services that are inaccessible to them. King County Metro (Washington State) provides such paratransit services through its Metro Access program. Metro Access has a direct interest in assessing pedestrian mobility along the PROW: (1) like other regional transit providers, they have a legal obligation to provide origin-to-destination paratransit rides to eligible individuals that cannot reasonably reach other means of transportation via the PROW, (2) they must determine eligibility for use of their services via local PROW surveys on a per-individual, and sometimes per-transit-request, basis, and (3) if possible, clients are preferentially directed to fixed transit through routes that are feasible and practical to the rider but may not necessarily be the most direct paths between origin and fixed-route stop. Since 2006 Metro Access staff, a team of pathway reviewers, has been slowly compiling a detailed database of pathways and barriers (including uneven terrain, elevation change greater than eight degrees, unmarked intersections, and improper/missing curb ramps). Over time, Metro Access has grown the effort to a dedicated team that collects the data in person for specific paths for frequent trips made by riders. The manual data collection consists of 1019 kilometers of sidewalks and crossings in King County. The team visits the origins and destinations of the rides in person and reviews the paths to bus stops for both legs of the trip. Included in the review are taking measurements and photographs of the sites. Eligibility reviews benefit both agencies and riders. Redirecting riders to the fixed route system frees up funds that can be put into other services
(including back into the paratransit system) and can provide a better experience to the individuals who request rides: the paratransit system must be booked days in advance and has large windows for pickup and drop-off, whereas fixed transit can often be independently navigated and may have more convenient time intervals. 

Until 2019, pathway reports collected by the pathway review team were not suitable to create pedestrian networks necessary for long-term needs assessment: reports were missing the vast majority of relevant public pedestrian pathways, potential obstacles and obstructions were stored as points separate from pathways (and therefore could not be automatically assessed via software without significant transformation and guesswork). Additionally, it was difficult to assess data for “staleness”, and it was collected in a purposefully incomplete and subjective manner: a pathway was reviewed primarily for the individual requesting paratransit services, so any information deemed irrelevant to that eligibility request was not collected. Much of the pathway review work was redundant in that the team collected and stored point of interest (POI) information readily available on other GIS services (OpenStreetMap, Google, etc). Finally, personally identifiable information (PII) could have been inferred from the data, presenting a privacy liability as well as a challenge to leverage sharing of data.

Metro Access is yet another user story demonstrating that comprehensively mapped information in the pedestrian environment is needed, and resonates concerns raised by other studies \cite{pearlman2014development, froehlich2019grand, bolten2017pedestrian}. The challenges faced by the Metro Access team were also prevalent with other stakeholders as described by \cite{bolten2022towards}. 
In general, surveying mapping tasks are laborious and infeasible if they are regularly collected manually. First, data collection and analysis are considerable because of the requirement for large amounts of data. Second, even once the data is collected, it cannot account for, nor be easily maintained through future changes to the environment (e.g. rerouting or reconstruction of the sidewalk). Presently, there are no devices for cities to collect this data in consistent and scalable ways, nor are open data standards being used to enable sharing with or across cities and organizations.

In recent years, Metro Access with partners, has been using an open data specification named OpenSidewalks \cite{opensidewalks} to perform collections of Geographic Information System (GIS) layers to store the results of these local surveys for future use: if there is already a known path or obstruction, a remote review can be performed to establish eligibility and any survey can focus on specific changes to infrastructure since the last review, rather than making redundant observations\cite{masstransit2021}. However, the general challenges of manual review remain. As pilot partners, we intended not only to address the technical challenges of automating the laborious parts of collecting OpenSidewalks transportation layer, and associated measurements, but also to improve the possibility of technology adoption through addressing organizational concerns around protecting personally identifiable information, as well as avoiding the costs and security concerns around large imagery storage and archival.



\begin{figure*}[t!]
  \centering
  \includegraphics[width=\linewidth]{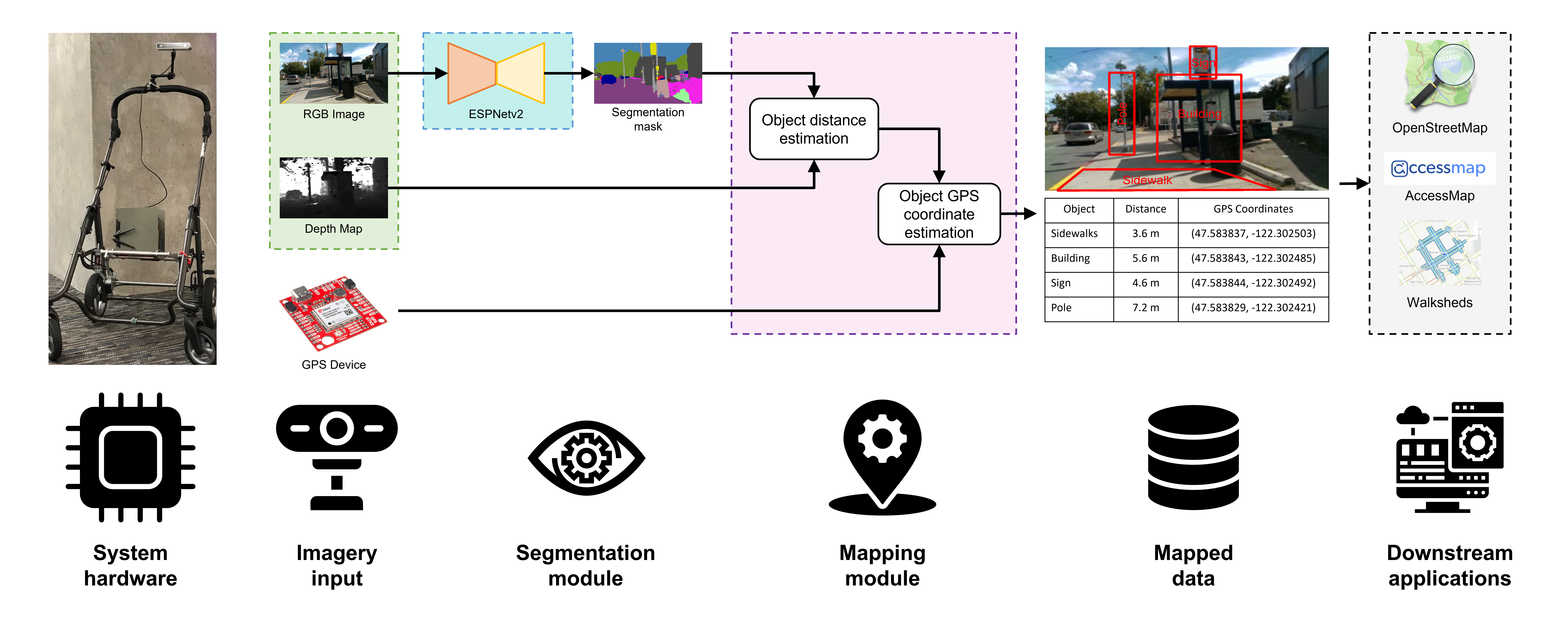}
  \caption{Overview of OASIS:  Geographic location of the objects in the scene and the connectivity of pedestrian path networks are inferred on-device with data from a stereo camera and a GPS device.}
  \label{fig:flow}
\end{figure*}

\section{Related work}
\label{sec:related_work}
In recent years, due in part to hardware improvements and the advent of computationally feasible Neural Networks, mapping with imagery data has become a rife area of research. This approach has gained primary prevalence in mapping automobile environments. some attempts in the field of pedestrian environment mapping have also been proposed and implemented.  

\paragraph{Semantic segmentation and automobile environment mapping} The automobile environment mapping has been widely studied, and semantic segmentation plays an important role in mapping automobile environments \cite{teichmann2018multinet, treml2016speeding, wu2022yolop, guo2021survey}. Several semantic segmentation architectures have been reported, including encoder-decoders \cite{badrinarayanan2017segnet, ronneberger2015u}, region-based \cite{dai2015convolutional, caesar2016region}, and cascade networks \cite{long2015fully}. Importantly, most of the semantic segmentation networks used in automobile mapping tasks are computationally expensive and are not suitable for mapping on edge devices. In this work, we extend efficient segmentation architecture, ESPNetv2 \cite{mehta2019espnetv2}, for the task of mapping sidewalks on the edge. As the base model, ESPNetv2 delivers similar performance while improving power efficiency over other state-of-the-art networks, including MobileNetv2 \cite{sandler2018mobilenetv2} and ShuffleNetv2 \cite{ma2018shufflenet}.

\paragraph{Pedestrian environment mapping} While the usefulness of segmentation models has been proven in the automobile environment, limited attempts have been made in the pedestrian environment. No widespread benchmark data sets include street-view images collected and annotated specifically for the pedestrian environment. Most of the related work on segmenting pedestrian paths focuses on using aerial satellite images \cite{senlet2012segmentation, ghilardi2016crosswalk}, or combining both satellite imagery and street-view imagery, but only applying to a specific single object class. For example, \cite{ahmetovic2017mind} utilizes both satellite images and Google Street View images to detect and localize street crossing. Those applications are often offline, primarily because they do not include any real-time information about the pedestrian environment. Other online works in the pedestrian environment focus on SLAM-based obstacle detection and avoidance \cite{morales20081km}, but such methods do not generate or store mapping information which can be used by downstream applications, for either pedestrian navigation (AccessMap \cite{accessmap, bolten2019accessmap}, SidewalkScore 
\cite{bolten2021towards}, or pathNav \cite{sinagra2019development}). To our knowledge, only one other hardware solution for pathways has been proposed, but it only addresses the standardization of surface integrity measures (i.e., measuring cracks and non-smooth surface disturbances), creates proprietary data, and does not provide an output pedestrian transportation graph to enable downstream routing applications \cite{smithsurface}. Other studies that map pedestrian path networks at scale are based solely on existing street (road) data. For example, Li et al. \cite{li2018semi} proposed a semi-automated method that uses roadway centerline data. The method generates sidewalks that should exist in theory, though manual editing is required.

\section{OASIS: An Open Automated Sidewalks Inspection System}
\label{sec:mapping_system}
In this section, we describe the Open Automated Sidewalks Inspection System (OASIS) for mapping sidewalks on portable devices commonly termed edge devices. OASIS, as shown in Figure \ref{fig:flow}, consists of two main modules: (1) segmentation module (Section \ref{sec:segmentation_module}) and (2) mapping module (Section \ref{sec:mapping_module}). The segmentation module classifies each pixel in an input image into semantic categories of interest. The mapping module takes the semantic segmentation mask, the depth map, and the GPS coordinates of the camera as an input and produces the GPS coordinates of each object present in the environment as an output. This not only helps us to understand the environment, but also allows us to register objects seen by the system to mapping libraries, such as OpenStreetMap (OSM), so that such information can be used in downstream applications. Figure \ref{fig:flow} shows the overall system and each component of OASIS is described in the following sub-sections.

\subsection{Segmentation module}
\label{sec:segmentation_module}
Semantic segmentation partitions a scene into several meaningful parts, decomposing it into its pixels and having each pixel labeled as belonging to a certain class of object. OASIS is built around the object classes that are meaningful in the context of outdoor pedestrian environment mapping. These classes are paths (road, sidewalk), and impassable obstacles which may be static (building, wall, fence, pole, traffic light, traffic sign, vegetation) or moving (person, rider, car, truck, bus, train, motorcycle, bicycle). Additionally, we segment sky and terrain to avoid false-positive object detection of street-side items in these areas. An integer ID is assigned to each pixel, representing its inclusion in the corresponding object class. 
The segmentation process uses an Image $I \in Z^{3 \times w \times h}$ as input, after the segmentation module assigns one of these integer IDs to each pixel in image $I$, a new image $S \in Z^{w \times h}$ called the image segmentation mask is produced.
Note that although OASIS is capable of segmenting objects among all the classes mentioned, the classes we choose for further analysis in this prototype system are specific to those static objects in the pedestrian environments: pole, traffic light, traffic sign, and building, which are commonly presented in the street environment and typically hard or too frequent to map each instance by hand. 
As discussed in Section \ref{sec:related_work}, the state-of-the-art segmentation models are the Convolutional Neural Networks (CNN). We choose ESPNetv2 \cite{mehta2019espnetv2} as the base segmentation model in our prototype system. ESPNetv2 greatly reduced the computational power needed while maintaining comparable results to other state-of-the-art models, which is important to deploy the model on an edge device. Each RGB image $I$ captured by the stereo cameras is used as input to ESPNetv2, generating the resulting segmentation mask $S$. Each pixel in the segmentation mask has an integer value representing the class that has the highest probability among all classes as predicted by the model.

Though the CNN models have high reported segmentation accuracy in outdoor environments, the models for the purpose of outdoor mapping have generally been trained on imagery acquired from a vehicle dashboard, e.g. Cityscapes \cite{cordts2016cityscapes} and KITTI \cite{geiger2013vision}. The errors introduced by this dataset bias are non-negligible in our domain setting of pedestrian environment mapping. We use two approaches to resolve this problem. Firstly, our application aims to perform semantic segmentation while the camera is traveling on the sidewalks. This implies that the images captured and fed to OASIS are consecutively in time. Instead of separately using each frame in the image stream for inferring, we propose that utilizing the temporal information in consecutive frames helps improve inference results. We compute the union of masks from CNN across a number of temporally adjacent frames. This straightforward approach is shown to reduce over-segmentation errors and increase overall prediction accuracy.  Secondly, observing that sidewalk is the class that suffers the most from the change of viewpoint between training samples and testing samples, we retrain the model on the more diverse dataset coco-stuff \cite{caesar2018cvpr}, and fine-tune on a dataset containing our hand-labeled images captured in diverse sidewalk environments. Because a model trained with images captured from the car dashboard has an inherent bias that the sidewalk has a higher probability to be on the sides of the input image, a model trained on an automobile dataset often incorrectly predicts the majority of the sidewalk as an automobile road (see Figure \ref{fig:seg-qual} for an Example). This error is not acceptable in applications that require knowledge of the location and connectivity of sidewalks. After retraining on our pedestrian-centric dataset, the model prediction accuracy is greatly improved in real sidewalk environments. Evaluation of our segmentation module is discussed in Section \ref{sec:eval-seg-module}.

\subsection{Mapping module}
\label{sec:mapping_module}
In this section, we detail the design of our mapping module. With the output of the segmentation module, the mapping module tracks and registers each static object in the scene (Section \ref{sec:tracking}), and infers sidewalk walkability and connectivity by inferring the location and width of each sidewalk fragment (Section \ref{sec:mapping_sidewalk}).

\subsubsection{Tracking and registering static objects}
\label{sec:tracking}
After the segmentation mask $S$ of each image $I$ is generated, it is combined with its corresponding depth map to compute the distance from the objects to the system. A depth map is an image (array) in which each pixel contains information about the depth from each point in the 3D scene at that pixel to the camera, from the same viewpoint that the corresponding RGB image is captured, see inputs in Figure \ref{fig:flow} for an example of an RGB image and its corresponding depth map.

Using the segmentation mask, we found the pixels representing each segmented object in the scene. The arithmetic mean of the values of these pixels in the depth map is computed to represent the depth of the segmented object. After the distance information of each segmented object is obtained, this distance is combined with the system's GPS information to estimate the GPS location of each object in the current scene. The system extracts the latitude and longitude at the time that an image is taken, and then infers the relative direction of each object using its pixel coordinates and the heading of the camera. Using the GPS location of the starting point (system location), camera heading, and distance inferred, the GPS of each object seen by the camera can be estimated with the Haversine formula.

Since our goal is to infer the geographic information for the sidewalk environment, if the same static object appears in different frames, the system needs to be able to recognize it. In order to keep track of each unique object in consecutive frames, we implemented an object tracker using the centroid tracking method. Using the segmentation mask, the centroid (the geometric center)  of each segmented object is computed. When a new frame comes from the camera, the object tracker takes in a set of new centroids coordinates and polygons of the segmented objects, which are produced by the CNN model for the current frame. The object tracker tries to match new objects with the existing objects from previous frames. This is done by computing the Euclidean distance between each pair of new object centroids and the existing object centroids. Each new object will be associated with an existing object where the Euclidean distance is the smallest. If the number of existing objects is greater than the number of new objects, the existing objects that can not be matched with a new object will be marked as temporarily lost and they will be marked as permanently lost if they stay in the temporarily lost state for a number of frames that is greater than a set threshold (5 frames in our prototype system).  If the number of existing objects is less than the number of new objects, each new object that cannot be associated with an existing object will be assigned a new object ID.  Since we are tracking objects while the camera is moving, the pixel location of object centroids will change because of the movement of the scene even if the object itself is static. To incorporate this information, we compute the homography transformation between adjacent frames, which describes their spatial relationship in terms of rotation and translation. Then we use this transformation to warp the centroid of each object from the current frame to the next frame. This step brings each existing object centroid closer to the new object centroid and improves the object tracking process. Section \ref{sec:eval-mapping-module} evaluates the mapping results for static objects.

\subsubsection{Mapping sidewalks}
\label{sec:mapping_sidewalk}
Different than the registration method we applied to the static objects, we use a different approach to map the sidewalk. Unlike other object classes, for a sidewalk, we not only want to know its geolocation, but the width of a sidewalk is also an important characteristic for measuring pedestrian path walkability. We did not attempt to track the sidewalks in sequential frames, instead, we directly compute the GPS coordinates of each sidewalk fragment with the Haversine formula. The width of each sidewalk fragment is then estimated with camera parameters through perspective projection.

As the geolocation and width of each sidewalk fragment being inferred, more importantly, the connectivity and walkability of a pedestrian path network are inferred. For inferring path connectivity, after the sidewalks for a neighborhood are inferred, a disconnection in the path network is identified when  any two nearest nodes in the predicted path network have a substantial gap in between. The inferred width can be used for evaluating sidewalks' accessibility and walkability. For example, sidewalks with insufficient width for certain mobility devices (e.g. wheelchairs) to pass will be marked as inaccessible to those users.

\section{Hardware setup}
\label{sec:hardware}
\begin{figure}[b!]
  \centering
  \includegraphics[width=\linewidth]{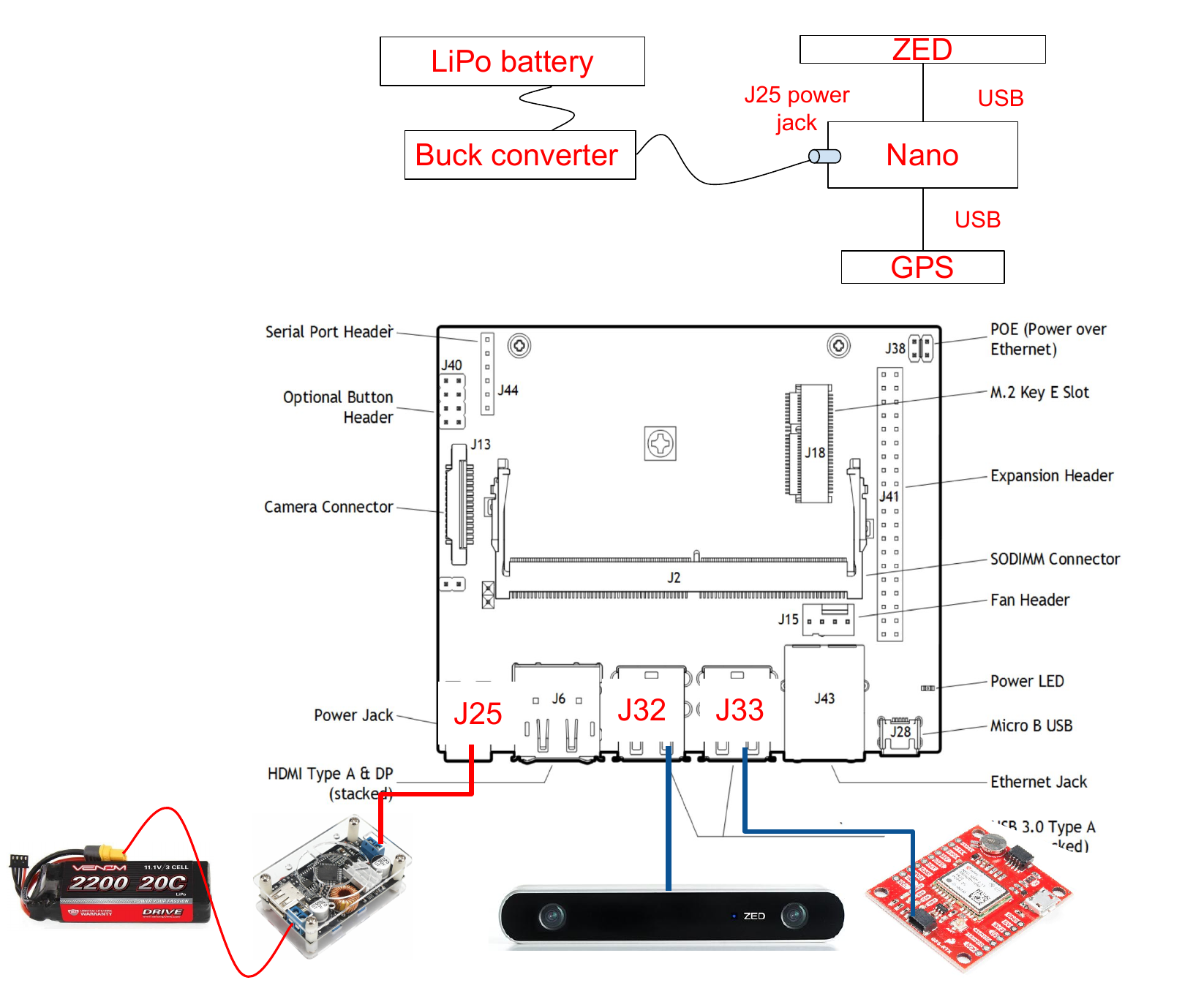}
  \caption{System schematic of OASIS}
  \label{fig:schematic}
\end{figure}

In this section, we describe the hardware setup of our prototype system used in our pilot study (Section \ref{sec:results}).  The schematic illustrating the overall hardware setup is shown in Figure \ref{fig:schematic}. The system consists of three hardware modules: (1) imaging module (Section \ref{sec:imaging_hardware}), (2) GPS module (Section \ref{sec:imaging_hardware}), and (3) computation module (Section \ref{sec:computation_hardware}). The imagery data from the imaging module and the GPS data (synchronized with each image) from the GPS module are processed at the computation module to produce mapping results.

\subsection{Imaging module}
\label{sec:imaging_hardware}
We use a stereo camera and a three-axis camera jig as the imaging module. The stereo camera is used to capture the RGB images and the corresponding depth image (shown in Figure \ref{fig:flow}), which are necessary for inferring the geolocation of infrastructures in the pedestrian environment and sidewalks connectivity. Based on the study comparing the performance of different stereo cameras in practical settings \cite{zhang2019stereo}, we use a ZED stereo camera (ZED) for better performance in diverse environments. The ZED stereo camera is a passive stereo vision camera that uses two lenses on each side of it to capture the scene. It then finds the matching points between the left view and right view of the same scene, and uses stereo vision algorithms to compute the depth of each sensed object through triangulation methods.

The ZED stereo camera is mounted onto the collection jig parallel to the ground. The spatial resolution was set to $1280 \times 720$ pixels for both the RGB images and the depth maps. The frame rate was set to 15 frames per second. The Field of View (FOV) of the ZED camera is $90^{\circ} \times  60^{\circ}$ (Horizontal x Vertical). The camera is calibrated per manufacturer specifications with tools provided by the manufacturers before collecting data.

\subsection{GPS module}
\label{sec:gps_hardware}
A GPS receiver is used to record the GPS information while the images are taken. This information, in conjunction with the imagery data, is used to infer the geolocation of each object of interest. The receiver in our prototype system is a u-blox ZED-F9P module (F9P). It has an accuracy of $\pm 0.75$ meters when tested in outdoor pedestrian environments. The GPS location of the system is recorded every 0.5 seconds using data from the GPS receiver. At this recording rate, while the system travels in the sidewalks at normal speeds, when each captured image is synchronized with the GPS data, it has a location error of $< 0.7$ meters, providing sufficient resolution for mapping. 

\subsection{Computation module}
\label{sec:computation_hardware}
The computing device is used for on-board semantic segmentation and geographic information inference. Because NVIDIA's GPU is required to fully utilize the Software Development Kit (SDK) of the ZED camera, and to optimally run CNN-based segmentation models, we use NVIDIA Jetson Nano which operates at under 10 watts as the computing device. With this device, we are able to map the sidewalks on portable edge devices.


\section{Pilot study and experiments}
\label{sec:results}
This section describes the pilot study we conducted with OASIS. The study consists of two parts. Section \ref{sec:quan hardware} provides an evaluation of OASIS at the hardware level. In Section \ref{sec:eval_mapping}, we use OASIS to map 3 neighborhoods and compare its mapping outcomes to the ones generated by human surveyors from local paratransit pathway review teams.

\subsection{Analysis of OASIS at hardware-level}
\label{sec:quan hardware}
\begin{figure}[t!]
  \centering
  \includegraphics[width=\linewidth]{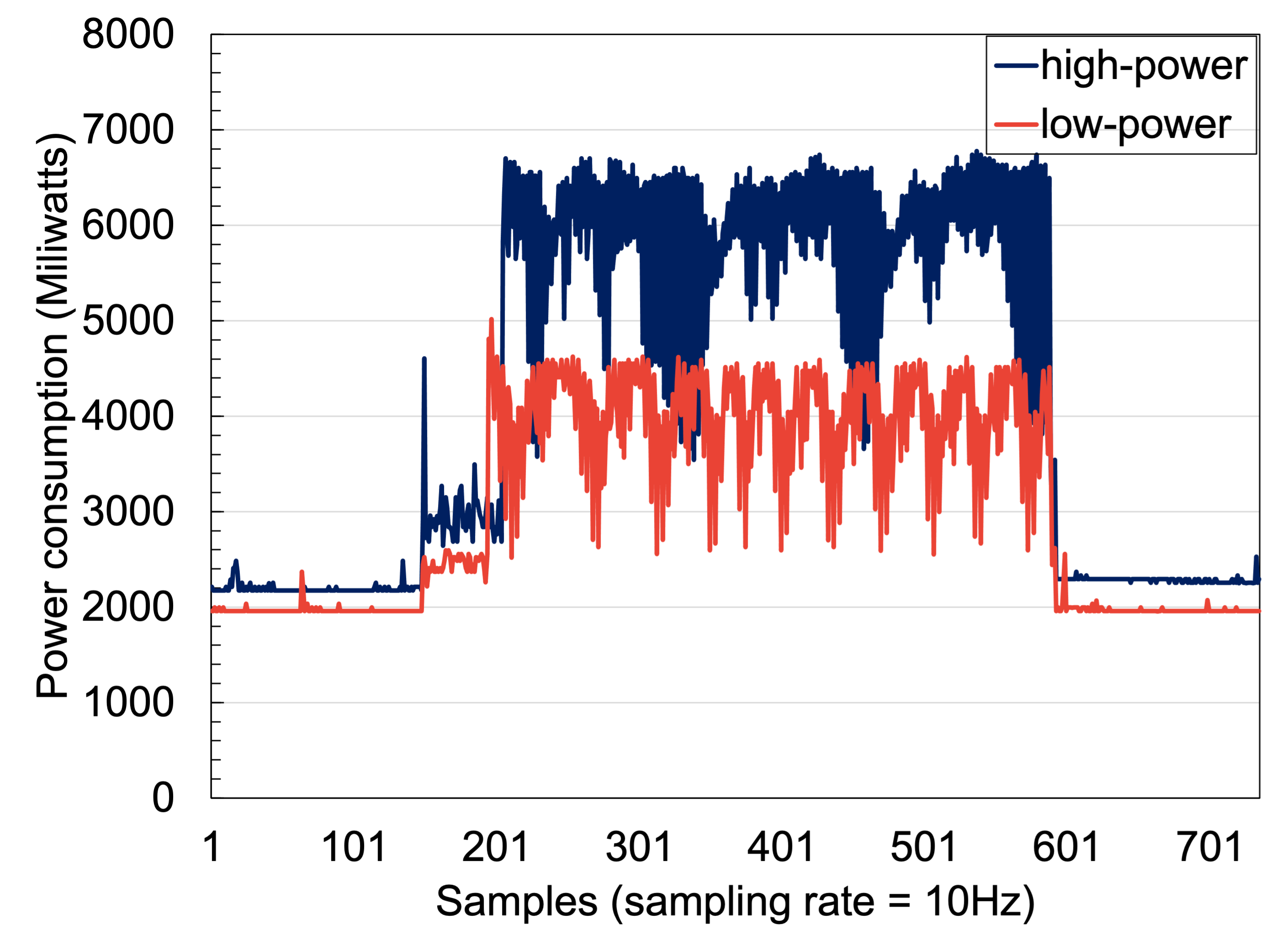}
  \caption{Power consumption of OASIS when mapping in two different power modes}
  \label{fig:power}
\end{figure}

The system performance is tested when OASIS runs continuous mapping in the pedestrian environment in two power modes. OASIS can operate at two power modes: (1) low-power (max. power: 5 W) and (2) high-power (max. power: 10 W). We evaluated OASIS under both power modes and measured different hardware metrics, including power consumption, memory usage, CPU usage, and GPU usage, for hardware-level analysis. Figure \ref{fig:power} shows the power consumption changes over time when OASIS runs at the two different power modes, also comparing to the base power consumption when OASIS is idle in each mode. Table \ref{tab:resource} shows the average resource usage of OASIS for a duration of 5 minutes during which the system is constantly running and generating mapping results. Overall, these metrics, especially low power consumption (about 4.5 watts and 6.5 watts in low-power and high-power mode respectively), suggest that OASIS can be integrated with a powered mobility device (e.g. a wheelchair or scooter) easily to power downstream applications. 

\begin{table}[t!]
\centering
\caption{Average resources usage on OASIS during mapping}
\label{tab:resource}
\resizebox{\columnwidth}{!}{
\begin{tabular}{lcccc} 
    \toprule[1.5pt]
   \textbf{Power mode} & \textbf{Avg. power consumption} & \textbf{RAM usage} & \textbf{GPU utilization} & \textbf{CPU utilization} \\
 \midrule
 \textbf{Low (5w)} & 3.98 W & 3006 MB & $82\%$ & $11.4\%@921MHz$ \\
 \textbf{High (10w)} & 5.87 W & 3058 MB & $84\%$ & $10.5\%@1479MHz$ \\
  \bottomrule[1.5pt]
\end{tabular}
}
\end{table}

\subsection{Analysis of pedestrian environment mapping with OASIS}
\label{sec:eval_mapping}
In this section, we describe our study for mapping sidewalks and infrastructures in real pedestrian environments and compare our mapping outcome to the ones generated by human surveyors from the local paratransit pathway review team. The studies were conducted in three geographically different areas: Bellevue, Redmond, and Seattle.

\begin{figure*}[tbh]
  \centering
  \begin{subfigure}{2\columnwidth}
  \centering
  \includegraphics[width=0.3\linewidth]{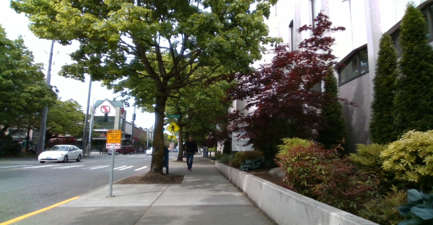}
  \includegraphics[width=0.3\linewidth]{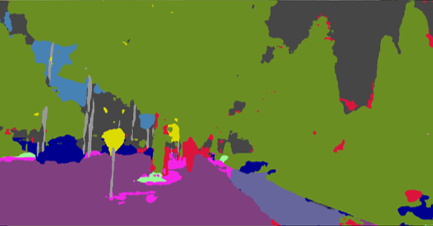}
  \includegraphics[width=0.3\linewidth]{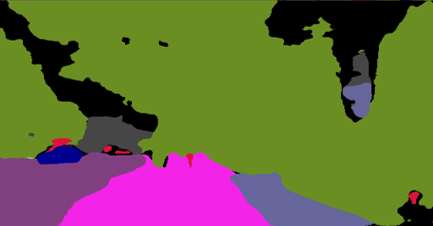}
  \end{subfigure}

  \begin{subfigure}{2\columnwidth}
  \centering
  \includegraphics[width=0.3\linewidth]{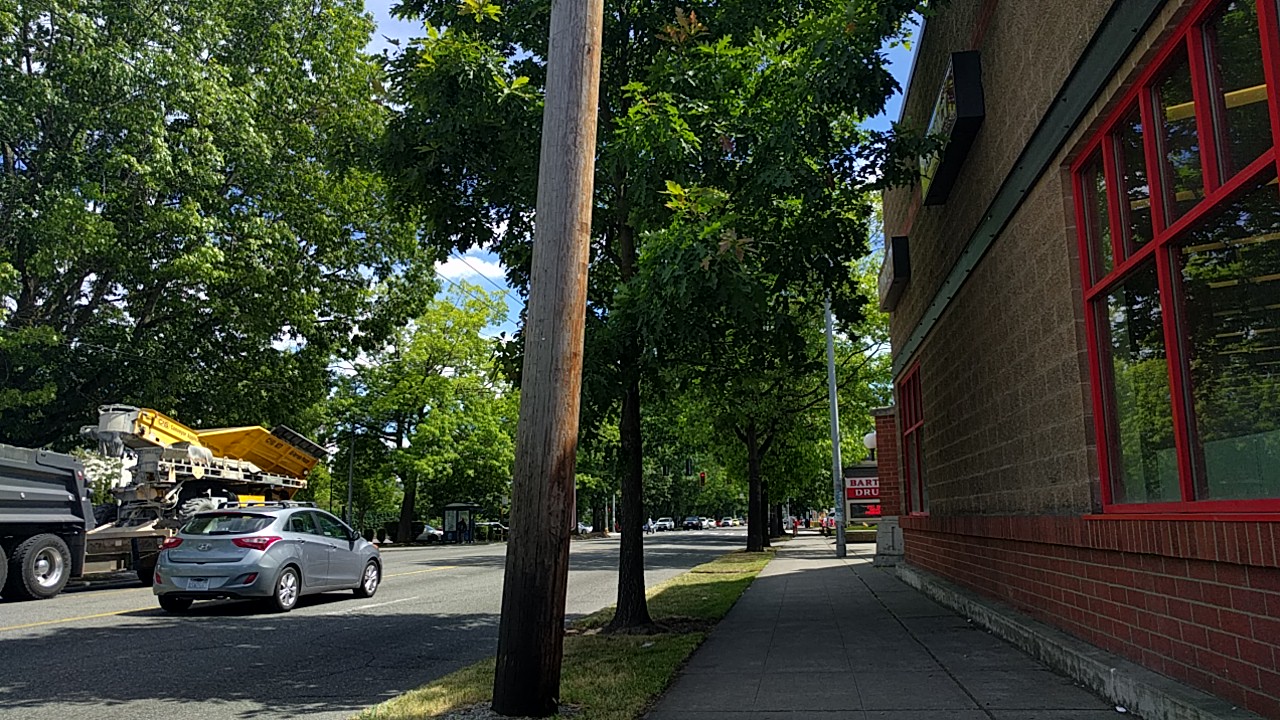}
  \includegraphics[width=0.3\linewidth]{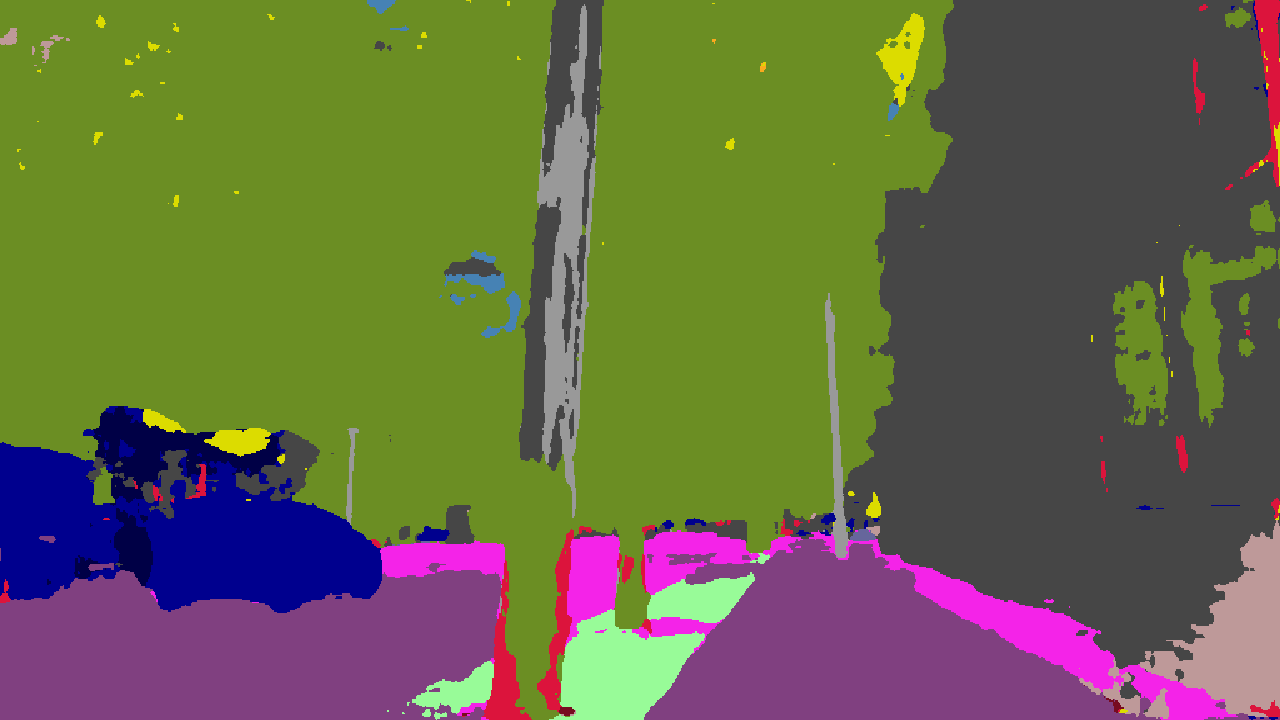}
  \includegraphics[width=0.3\linewidth]{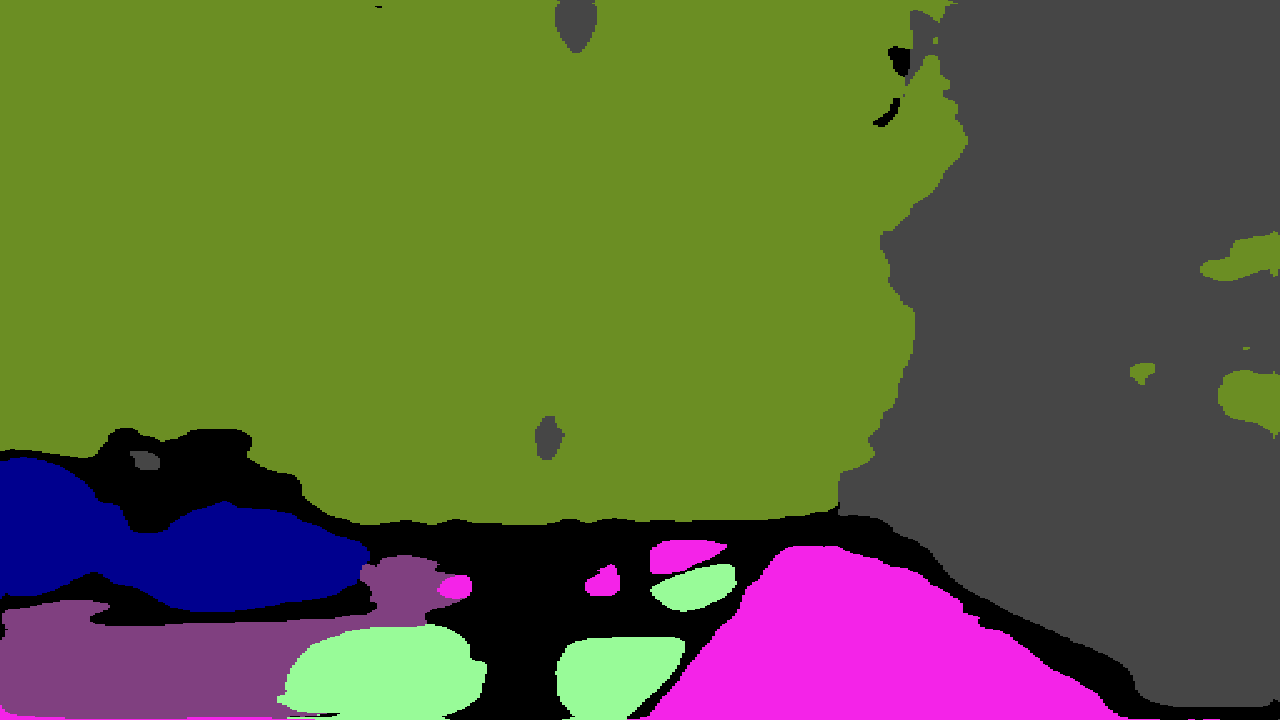}
  \end{subfigure}
  
   \begin{subfigure}{2\columnwidth}
  \centering
  \includegraphics[width=0.3\linewidth]{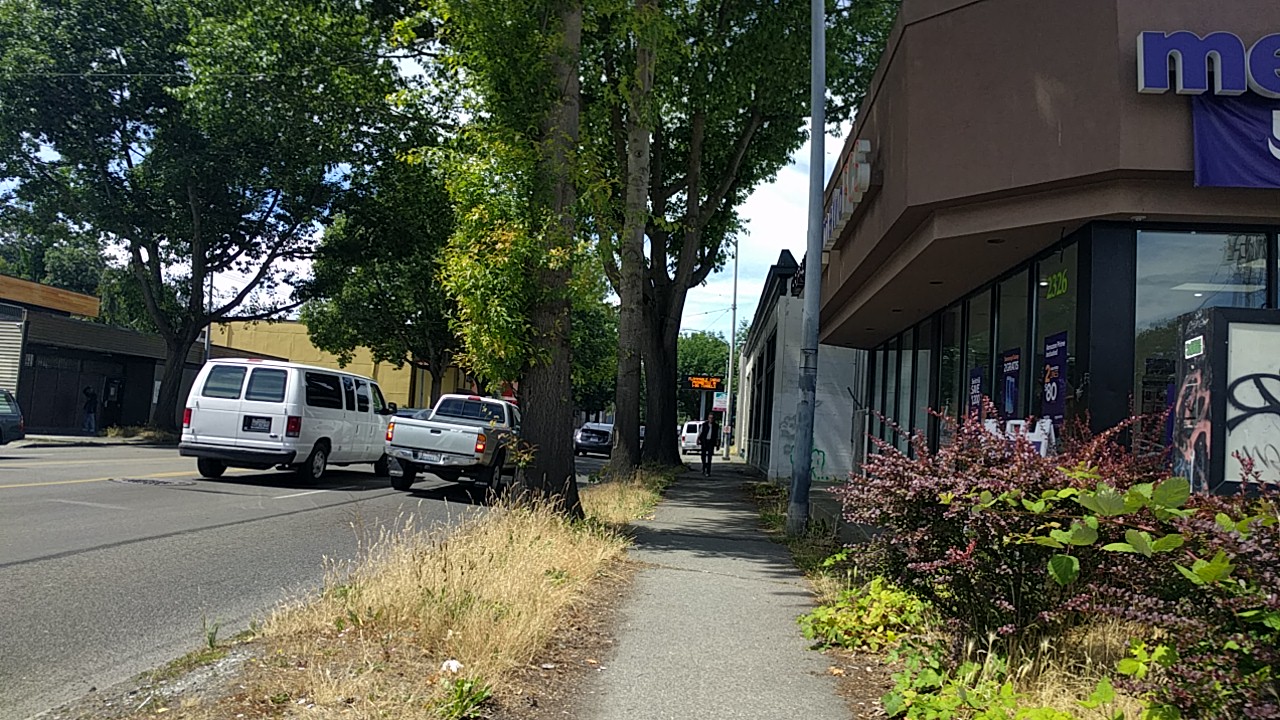}
  \includegraphics[width=0.3\linewidth]{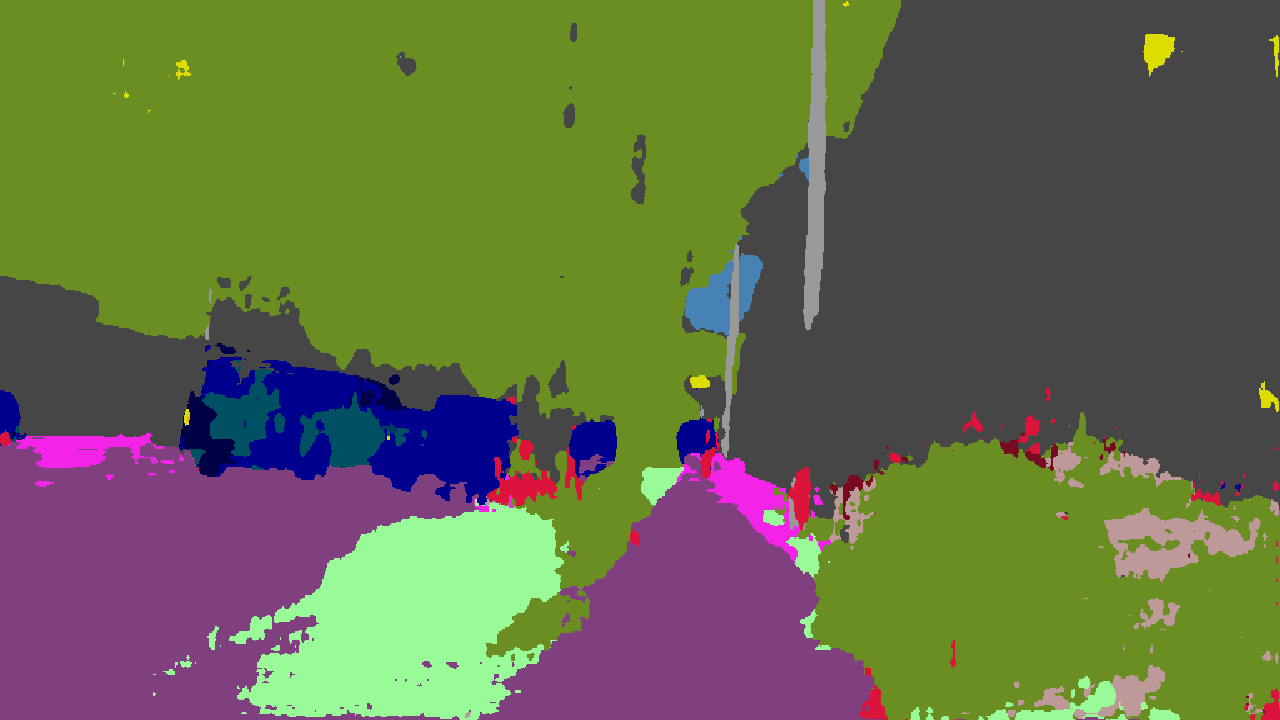}
  \includegraphics[width=0.3\linewidth]{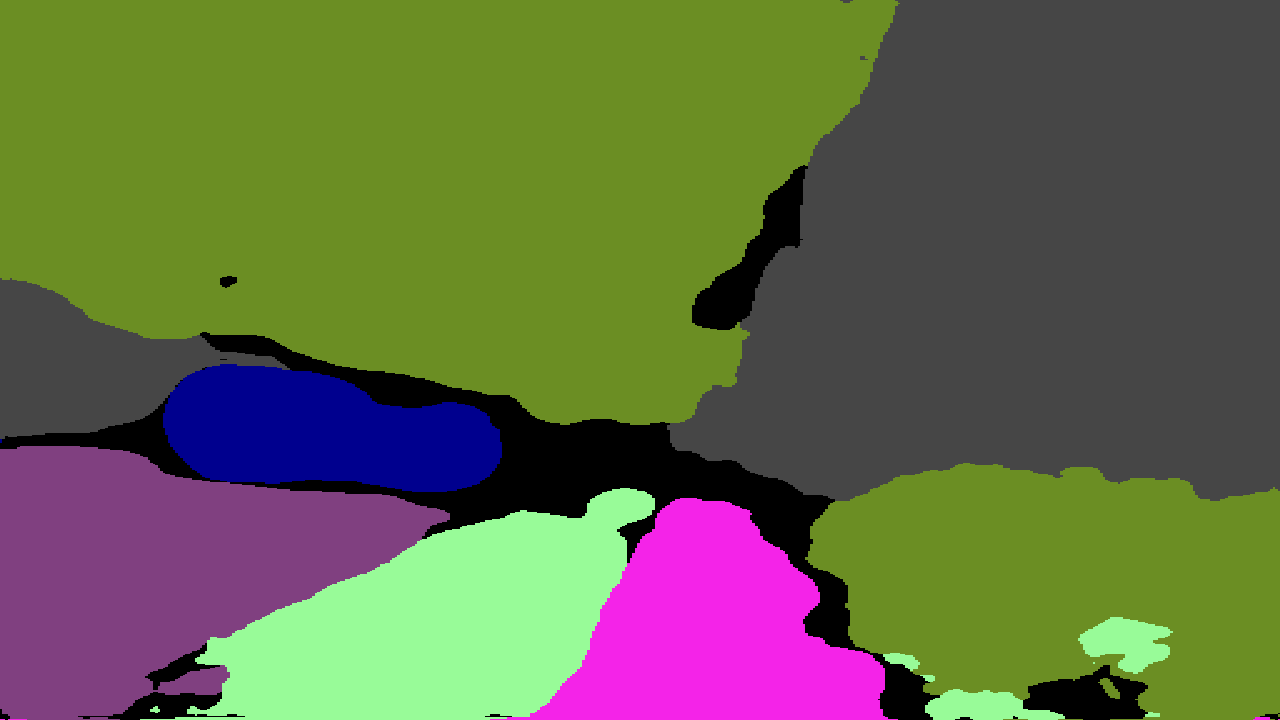}
  \end{subfigure}
  
 \begin{subfigure}{2\columnwidth}
  \centering
  \includegraphics[width=0.3\linewidth]{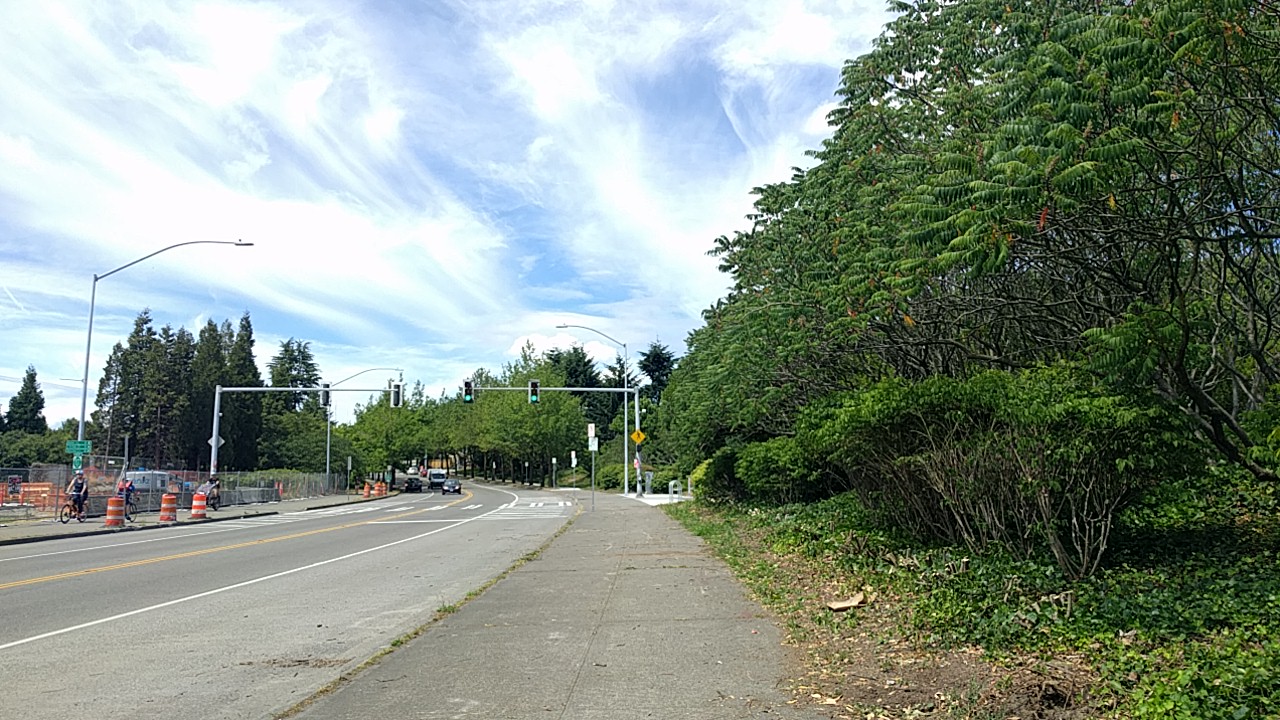}
  \includegraphics[width=0.3\linewidth]{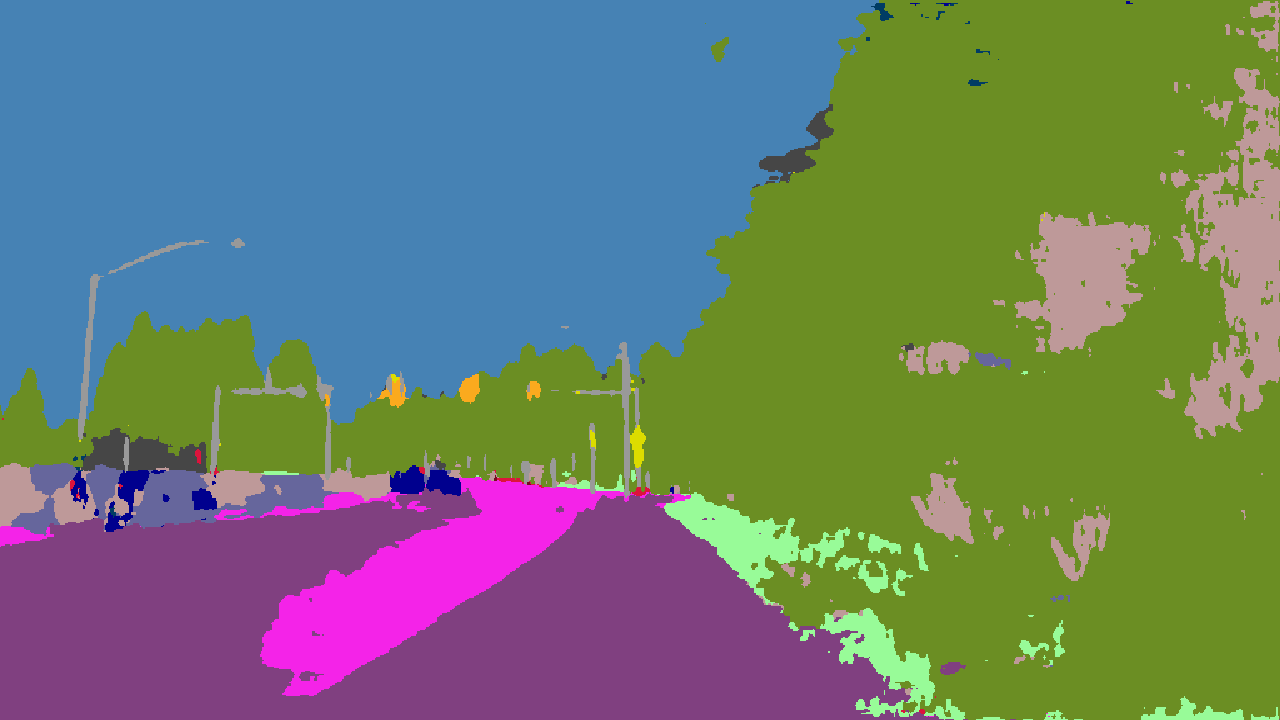}
  \includegraphics[width=0.3\linewidth]{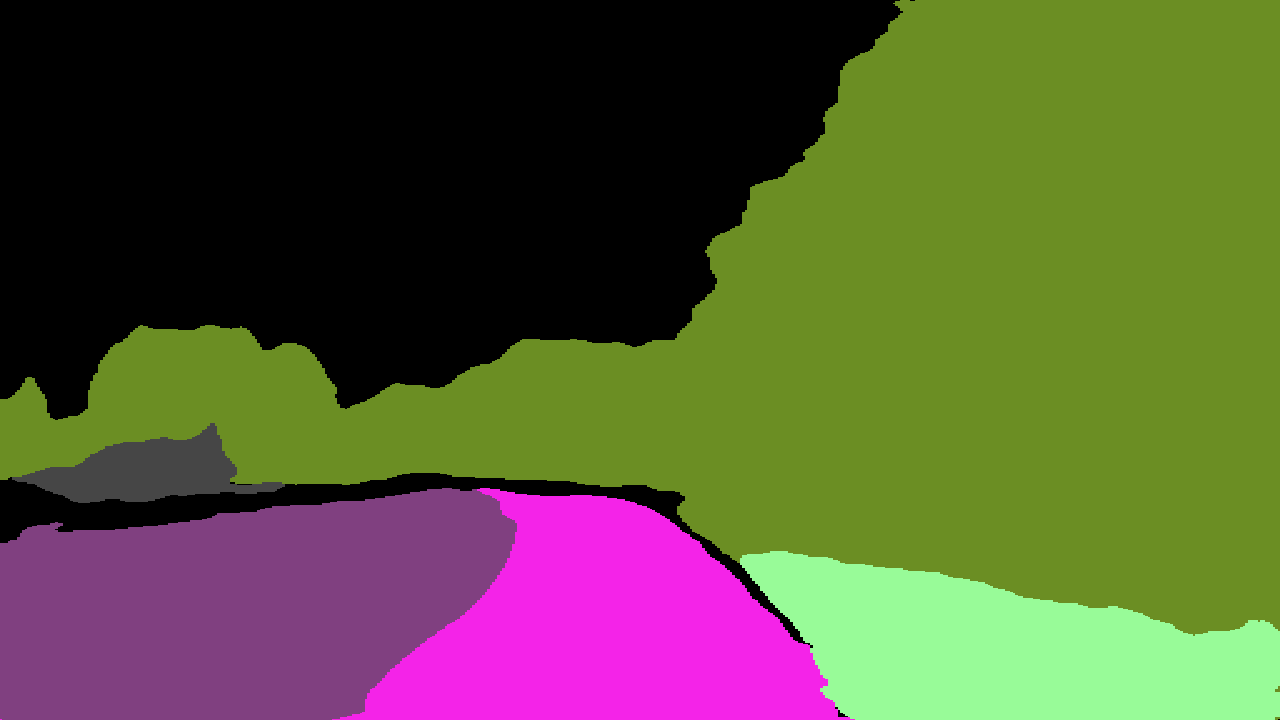}
  \end{subfigure}
  \caption{Qualitative performance of the segmentation module. \textcolor{blue}{Left:} Input image. \textcolor{blue}{Middle:} Segmentation masks produced by ESPNetv2. \textcolor{blue}{Right:} Segmentation masks produced by ESPNetv2-PED. In the segmentation masks, the sidewalk is represented in \colorbox{megenta}{magenta}, the automobile road is represented in \colorbox{darkpurple}{dark purple}, and the terrain is represented in \colorbox{lightgreen}{light green}.}
  \label{fig:seg-qual}
\end{figure*}

\begin{table*}[tbh]
\centering
\caption{Performance of the segmentation module in the pedestrian environment}
\label{tab:object_accuracy}
\resizebox{2\columnwidth}{!}{
\begin{tabular}{ lcccccccccccccccccccc } 
    \toprule[1.5pt]
  & \multicolumn{3}{c}{\textbf{Building}} && \multicolumn{3}{c}{\textbf{Pole}} && \multicolumn{3}{c}{\textbf{Traffic Light}} && \multicolumn{3}{c}{\textbf{Traffic Sign}} && \multicolumn{3}{c}{\textbf{Average}} \\
  & \textbf{IoU} & \textbf{Precision} & \textbf{Recall} && \textbf{IoU} & \textbf{Precision} & \textbf{Recall} && \textbf{IoU} & \textbf{Precision} & \textbf{Recall} &&  \textbf{IoU} & \textbf{Precision} & \textbf{Recall} && \textbf{IoU} & \textbf{Precision} & \textbf{Recall}\\
  \cmidrule[1pt]{2-4} \cmidrule[1pt]{6-8} \cmidrule[1pt]{10-12} \cmidrule[1pt]{14-16} \cmidrule[1pt]{18-20}
  \textbf{ESPNetV2} & 0.815 & 0.654 & 0.895 && 0.441 & 0.841 & 0.659 && 0.423 & 0.577 & 0.645 && 0.539 & 0.781 & 0.618 && 0.547 & 0.713 & 0.704 \\
  \textbf{ESPNetV2-PED} & 0.869 & 0.691 & 0.932 && 0.502 & 0.882 & 0.721 && 0.485 & 0.613 & 0.713 && 0.572 & 0.815 & 0.686 && 0.607 & 0.750 & 0.763 \\
 
  \bottomrule[1.5pt]
\end{tabular}
}
\end{table*}

\begin{figure*}[tbh]
  \centering
  
  \begin{subfigure}{1.9\columnwidth}
   \centering
  \includegraphics[width=0.3\columnwidth,height=85px]{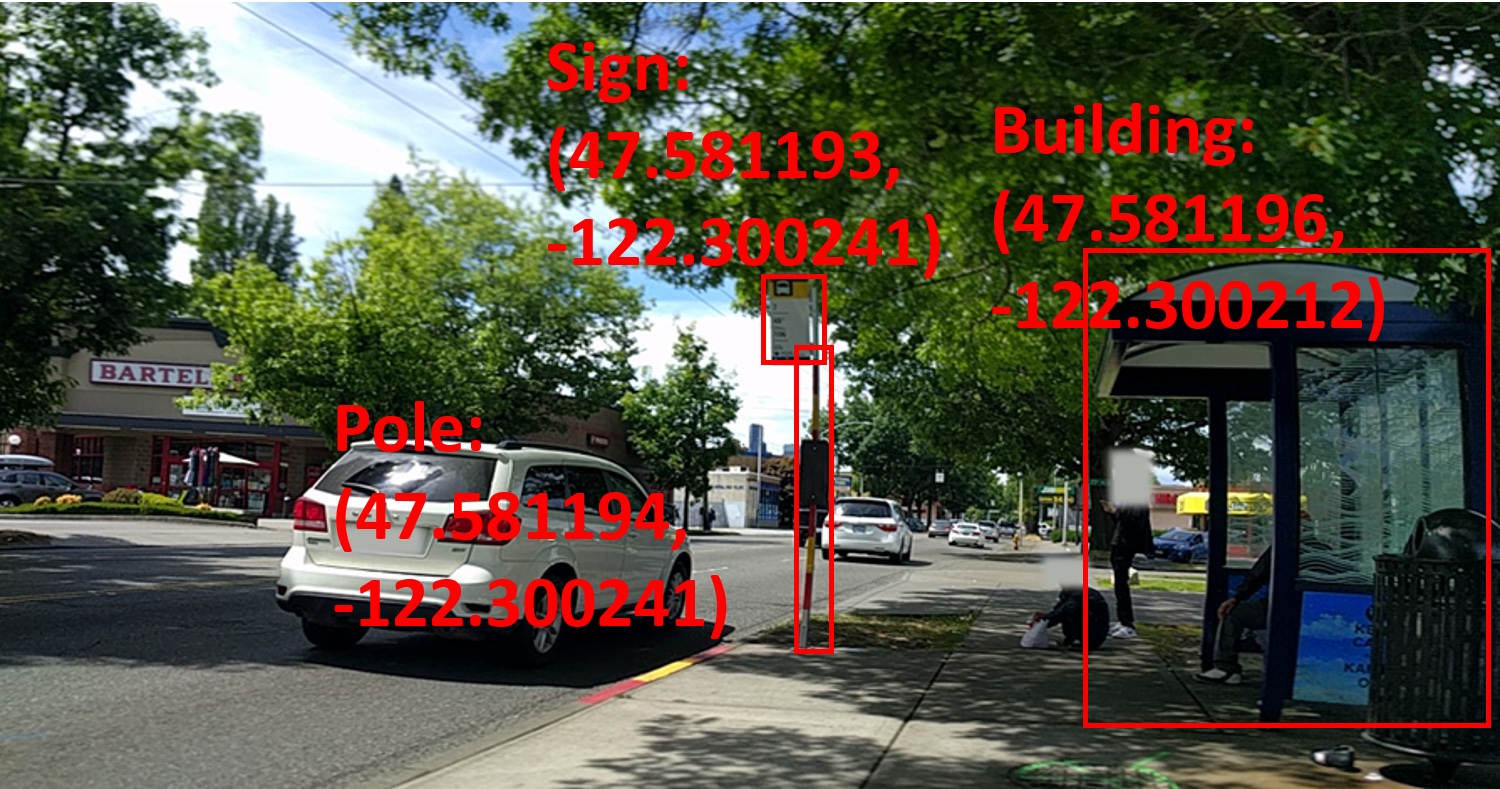}
  \includegraphics[width=0.3\columnwidth,height=85px]{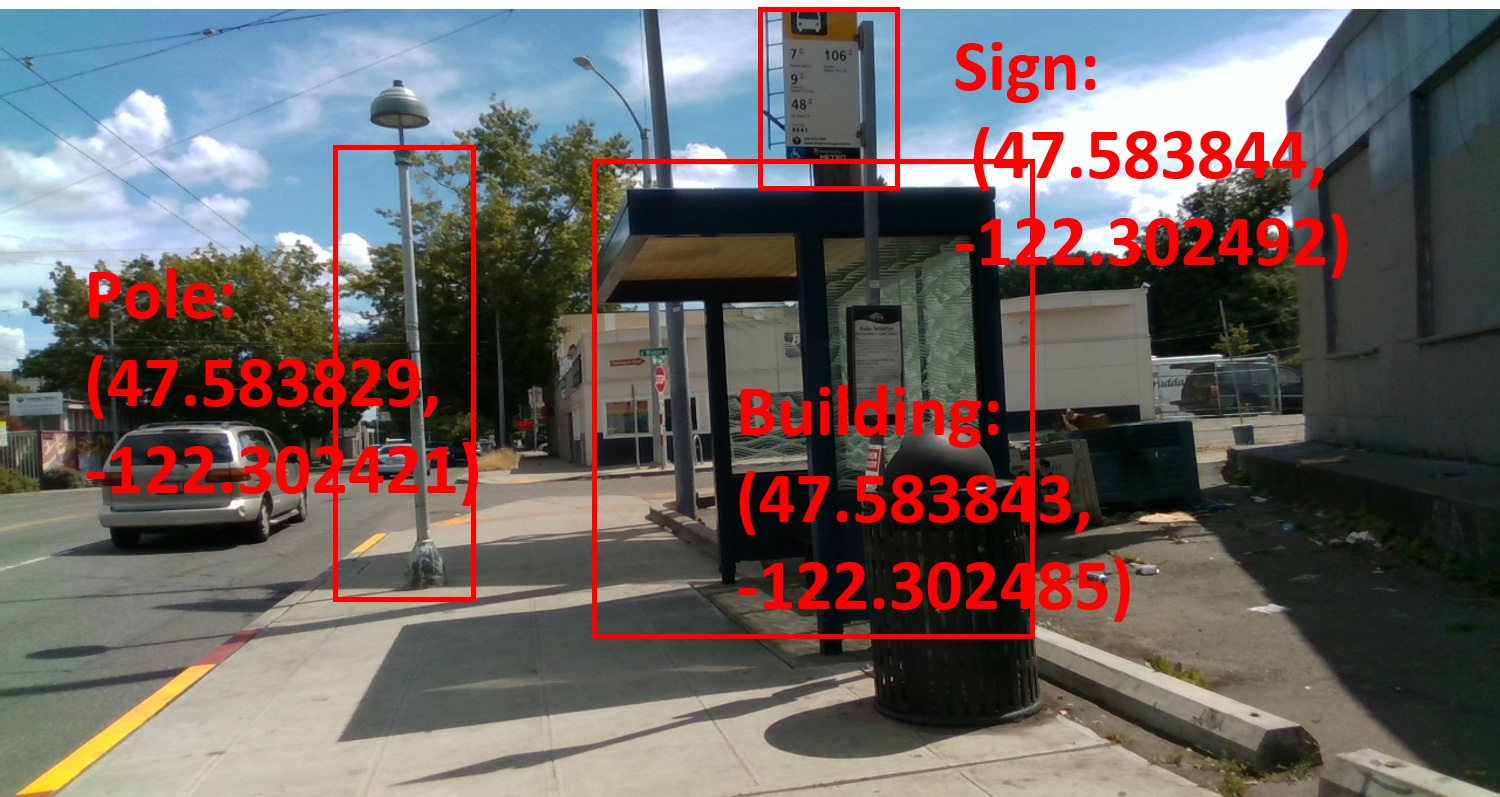}
  \includegraphics[width=0.3\columnwidth,height=85px]{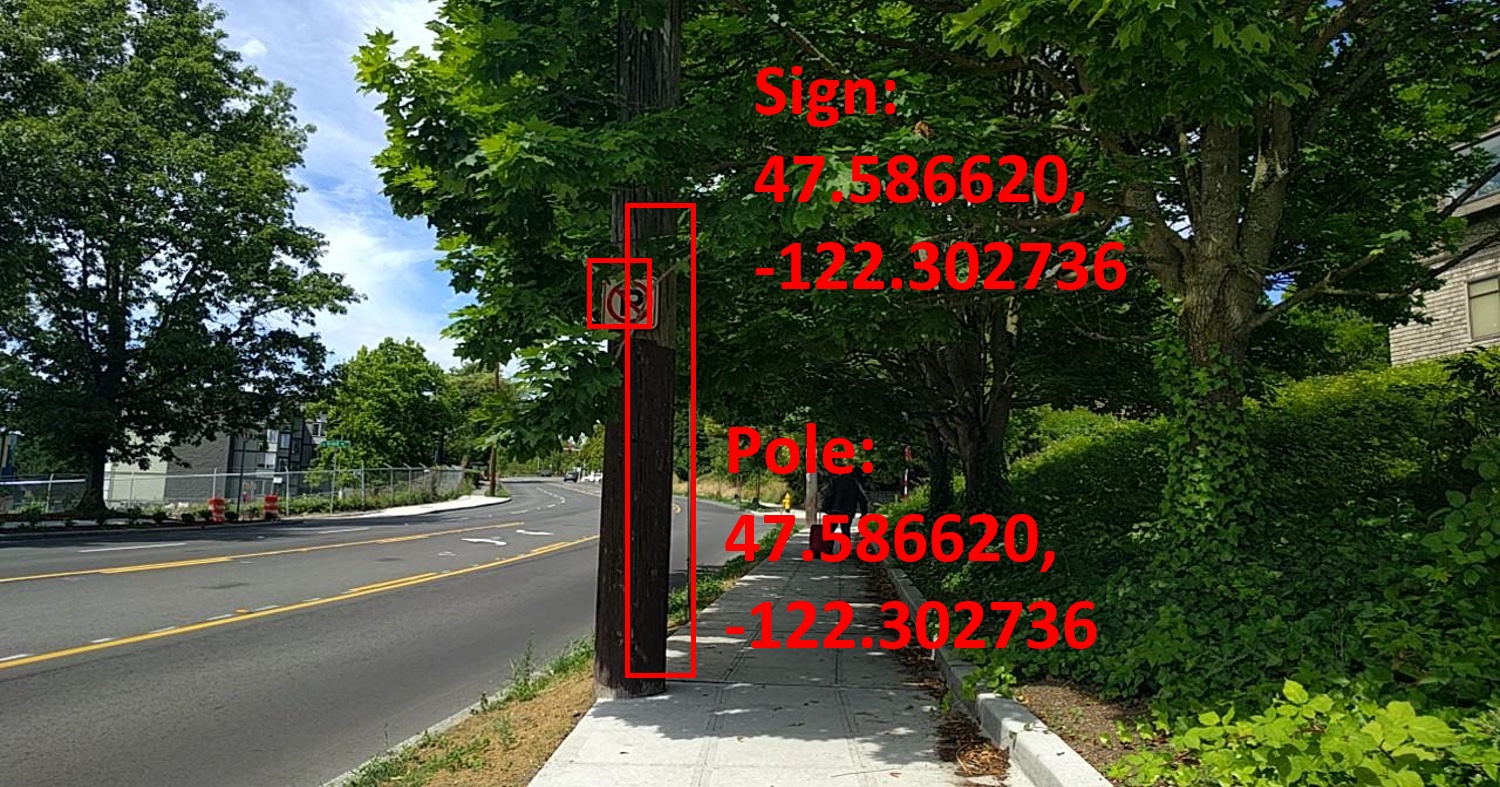}
  \caption{Mapped infrastructures}
  \label{fig:infras}
\end{subfigure}
   
\begin{subfigure}{1.9\columnwidth}
  \centering
  \includegraphics[width=0.3\columnwidth,height=85px]{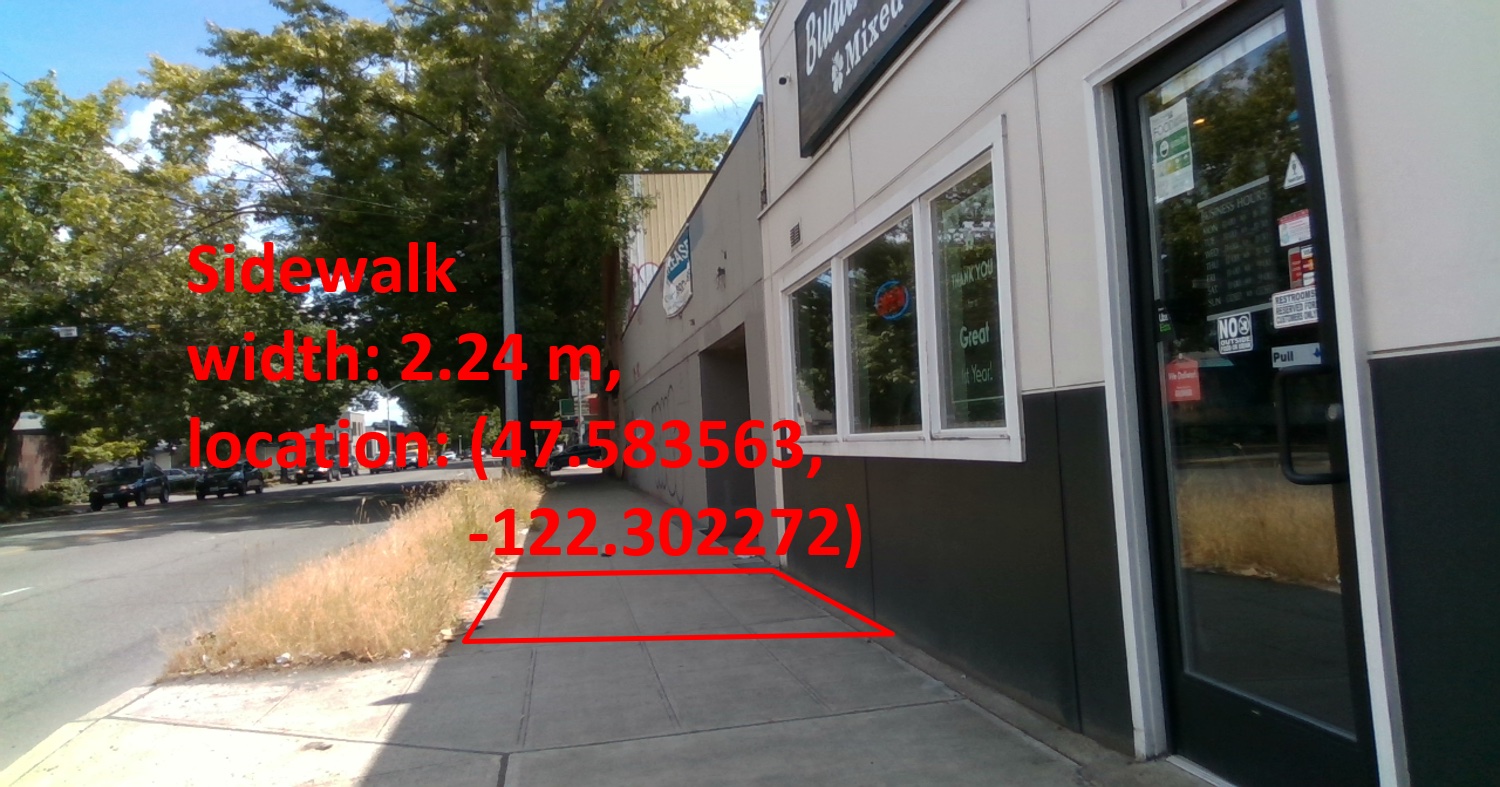}
  \includegraphics[width=0.3\columnwidth,height=85px]{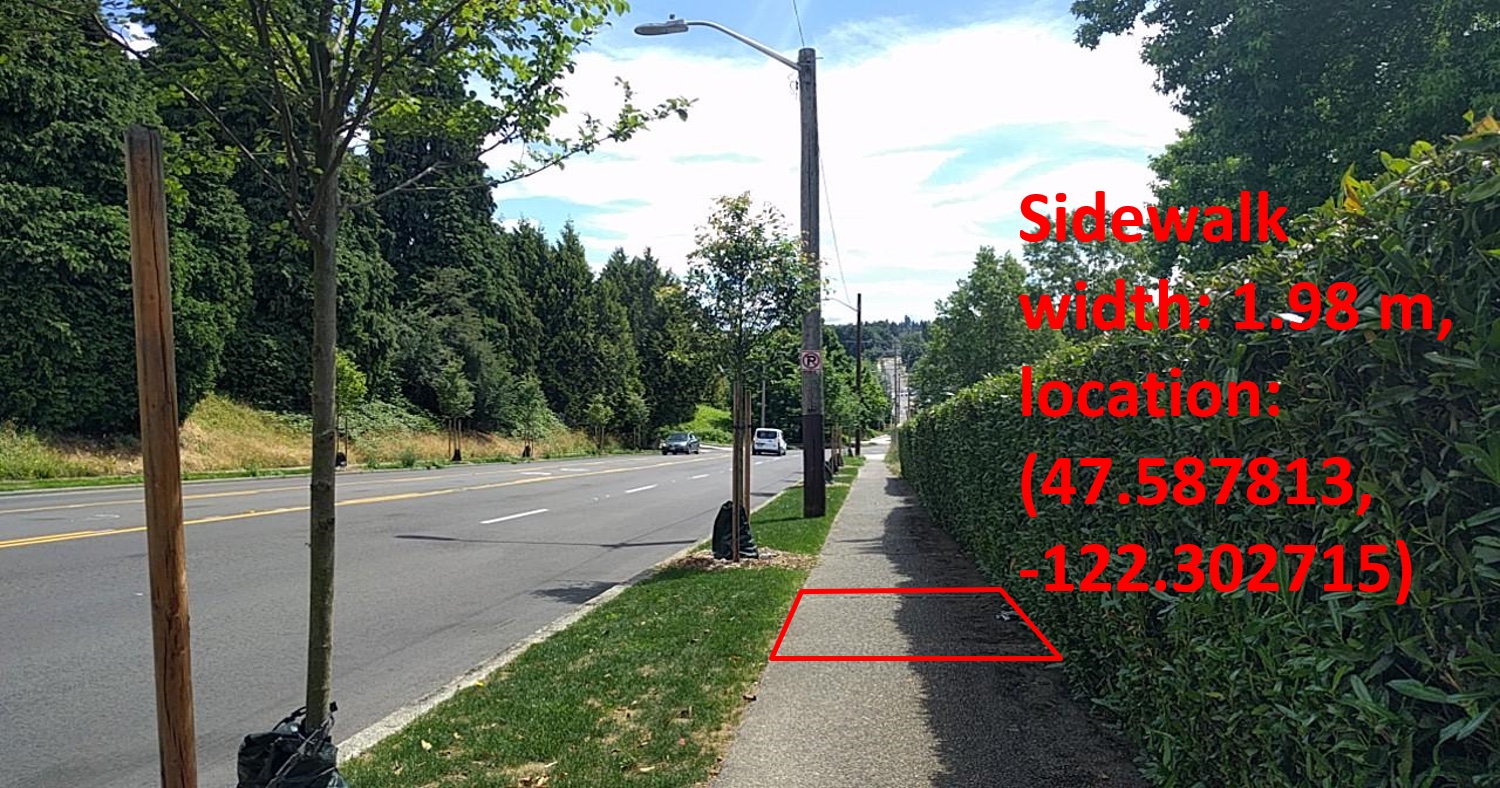}
  \includegraphics[width=0.3\columnwidth,height=85px]{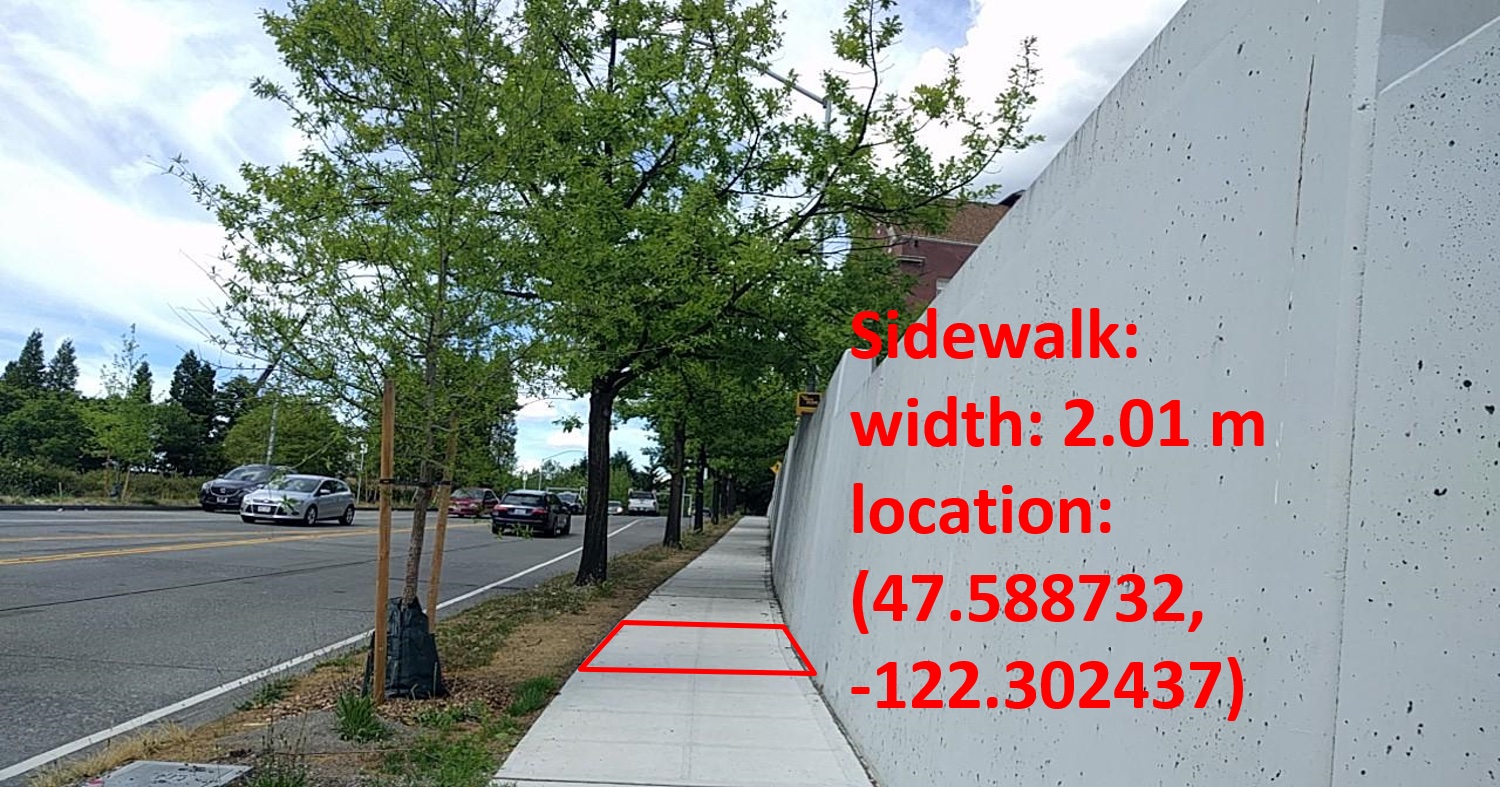}
  \caption{Mapped sidewalks}
  \label{fig:sidewalk}
   \end{subfigure}

\caption{Qualitative mapping results: (a) The predicted geolocations of the infrastructures on the sidewalks. (b) The predicted geolocations of sidewalks and the inferred width of each sidewalk fragment.}
\label{fig:qualitative results}
\end{figure*}

\subsubsection{Segmentation module}
\label{sec:eval-seg-module}
We start by evaluating the performance of the segmentation module in the pedestrian environment by comparing different segmentation models. ESPNetv2 trained on the commonly used Cityscapes dataset is used as the baseline, and for comparison, ESPNetv2-PED denotes the model we trained for the pedestrian environment with the method described in Section \ref{sec:segmentation_module}. The models' performance is visualized in Figure \ref{fig:seg-qual}. The middle column shows the qualitative segmentation results from ESPNetv2 and the right column shows the segmentation result from ESPNetv2-PED. With ESPNetv2, because of the inherent bias that the training data is captured from a car dashboard, this model incorrectly predicts the majority of sidewalks to be the automobile road, making it impossible to generate correct sidewalk mapping information. On the other hand, with ESPNetv2-PED, we can see that the model correctly segments the sidewalk after incorporating pedestrian-centric data in training, enabling us to correctly infer sidewalk mapping information. Quantitatively, we use three standard metrics to evaluate the segmentation module: (1) pixel-wise intersection over union (IoU), (2) precision, and (3) recall of our objects of interest. As shown in Table \ref{tab:object_accuracy}, high precision and recall on the common objects demonstrate the system successfully detects objects in real pedestrian environments. We also observe that ESPNetv2-PED outperforms the baseline model in the pedestrian environment in every category.

\subsubsection{Mapping module}
\label{sec:eval-mapping-module}
While the system travels in the pedestrian pathways, street-side images are captured by the camera and processed by the system to generate mapping predictions. Figure \ref{fig:infras} shows examples, where the infrastructures of pedestrians' interest are segmented by the system, and their distance and geolocations are inferred. On average, 62 obstacles that may impede pedestrians' travel are identified and located per kilometer of sidewalks. As an example, the bus station in Figure \ref{fig:infras} is predicted as a building and it will be treated as an obstacle in the navigation setting, similar to the situation where part of a building extends into the sidewalk and blocks the pedestrians' way. Besides mapping the infrastructures in the sidewalks, more importantly, we map the sidewalks to infer pedestrian path connectivity and walkability. Figure \ref{fig:sidewalk} shows examples of the mapped sidewalk fragments, with their geolocation and width inferred. Using the inferred mapping of these fragments, a pedestrian path network is generated. 

To quantitatively evaluate the mapping module, we compare the predicted geolocation of each static object to its real geolocation in the frames that were manually annotated. It is to be noted that the errors of distance estimation under various environmental conditions have been studied in our previous work \cite{zhang2019stereo}, hence we only measure the location error in this study. The location error of our mapped objects of interest (measured as the distance between two points in meters) is shown in Table \ref{tab:loc_error}. The low average error demonstrates the efficacy of OASIS in mapping the common infrastructures on sidewalks.  More importantly, to evaluate the predicted sidewalk network graph, we provide quantitative measures for both the predicted location and the predicted width for each of the three cities as shown in Table \ref{tab:sw_error}. The predictions are compared to the data collected and maintained by the city. Low RMSE demonstrates that the system effectively predicts the location and width of sidewalks. In addition, to evaluate the mapped sidewalks at the network graph level, we use metrics similar to the ones proposed by \cite{mattyus2017deeproadmapper} for road topology measurements, namely the precision, recall, and F1 score based on the assignments of predicted sidewalk edges (or crossing edges) to the corresponding edges in the ground truth graph. As shown in the last three columns in Table \ref{tab:sw_error}, our method maintains high precision and recall across three areas with significantly different built environments, demonstrating the mapping efficacy and accuracy of our method.

\begin{table}[tbh]
\centering
\caption{Error measure of mapped static objects locations}
\label{tab:loc_error}
\resizebox{0.95\columnwidth}{!}{
\begin{tabular}{ lccccc } 
    \toprule[1.5pt]
  & \textbf{Building} & \textbf{Pole} & \textbf{T light} & \textbf{T sign} & \textbf{Average}\\
 \midrule[1pt]
  \textbf{Mean Error (m)} & 0.511 & 0.620 & 0.424 & 0.441 & 0.496  \\
 \textbf{Standard Deviation Error (m)}  & 0.217 & 0.122 & 0.102 & 0.089 & 0.136 \\
  \textbf{Root Mean Squared Error (m)} & 0.554 & 0.632 & 0.436 & 0.449 & 0.514  \\
  \bottomrule[1.5pt]
\end{tabular}
}
\end{table}

\begin{table}[t]
\centering
\caption{Quantitative evaluation: sidewalks mapped by OASIS compared to human annotations}
\label{tab:sw_error}
\resizebox{1\columnwidth}{!}{
\begin{tabular}{lcccccccccccc} 
    \toprule[1.5pt]
    & \multicolumn{3}{c}{\bfseries Location error (m)} && \multicolumn{3}{c}{\bfseries Width error (m)} &&  \multirow{2}{*}{\textbf{Precision}} & \multirow{2}{*}{\textbf{Recall}} & \multirow{2}{*}{\textbf{F1}} \\
    \cmidrule[1pt]{2-4}\cmidrule[1pt]{6-8}
  & \textbf{Mean} & \textbf{STDEV} & \textbf{RMSE} && \textbf{Mean} & \textbf{STDEV} & \textbf{RMSE} \\
 \midrule[1pt]
  Redmond & 0.47 & 0.12 & 1.49 && 0.11 & 0.03 & 0.11 && 0.94 & 0.98 & 0.96\\
  Bellevue & 0.68 & 0.22 & 0.71 && 0.28 & 0.07 & 0.29 && 0.92 & 0.98 & 0.95\\
  Seattle & 0.53 & 0.11 & 0.54 && 0.14 & 0.04 & 0.15 && 0.96 & 0.99 & 0.97\\
  Average & 0.56 & 0.15 & 0.91 && 0.18 & 0.05 & 0.18 && 0.94 & 0.98 & 0.96\\
  \bottomrule[1.5pt]
\end{tabular}
}
\end{table}

\subsection{Towards efficient and scalable pathway review process with OASIS}
\label{sec:compare_paratransit}
One main consideration in our study was the feasible deployment and adoption of OASIS to assist pathway review teams in larger-scale applications. During our pilot study, we found OASIS has five benefits (1) While the system can be guided at a typical walking speed, the average time for mapping one mile of sidewalks is reduced six-fold (specifically, it took 120 min on average for an unassisted reviewer to collect one mile of pathways as opposed to 20 minutes with OASIS). OASIS can reduce the time and cost of collecting sidewalk mapping data by reducing the effort of human surveyors. (2) As discussed in Section \ref{sec:eval_mapping}, our method can provide accurate mapping information that is difficult to collect manually with consistency (e.g. width of sidewalks and every infrastructure in the scene is tedious to collect), in addition to providing accurate location and connectivity data of the sidewalks. (3) The data output is immediately available in an open and standardized format per Opensidewalk Schema specifications \cite{opensidewalksshcema}, resolving a common issue with denoising and postprocessing that the manual surveyor collection requires. This facilitates the data entry stage and improves the probability of data reuse and consumption by other stakeholders or downstream applications. (4) The output of OASIS is the mapped network data in the pedestrian layer, raw image data can be discarded for public privacy concerns. (5) Being built around portable edge devices, OASIS can be easily integrated into powered mobility devices and the data can be easily maintained up-to-date with minimal human intervention.
Though further adoption studies are required, the pilot findings suggest that OASIS can be adopted in audits and applications where fast and accurate assessment and re-assessment of sidewalks are needed. In addition, OASIS has minimal reliance on skilled surveyors once deployed at scale, making it suitable for larger-scale applications and organizational efforts. 

\section{Discussion and conclusion}
\label{sec:conclusion}
Novel technologies helping cities manage data pipelines about everything inclusive of its services and public infrastructure will be the driving force for future cities. Pedestrian ways, sidewalks, and footways are of primary concern for sustainable, accessible cities that are attempting to influence city inhabitants to make use of active transportation options rather than private vehicles. This paper proposes a novel urban sidewalk assessment approach using computer vision (machine learning) techniques on portable (edge) devices. 
We have implemented and experimented with OASIS and deployed it towards mapping sidewalk environments. OASIS enables private or public organizations to map sidewalks and their connectivity (in the form of a routable graph) quickly and with reasonable accuracy.

A primary contribution of this and similar technologies used for infrastructure mapping is in producing consistent, standardized data at scale. There is an opportunity for the field to contribute to an effective and resilient data ecosystem that will provide many stakeholders, including transportation agencies,  municipalities, public and private civic actors alike with relevant data to analyze, metricize, prioritize and possibly manage mobility and accessibility at city scale. Performing all computing on the edge may be attractive to municipalities where the citizenry would reject the use of cloud infrastructure to obtain and potentially keep sidewalk imagery that they might deem as an invasion of privacy or security. The availability of the data on a portable edge device platform opens further opportunities to inform (via a map) automated applications that navigate in sidewalk environments (such as autonomous wheelchairs and self-driving delivery robots). The output mapping results that are open and shared are a crucial step towards this vision since they are not reliant on any proprietary street-side imagery contributions. Specifically, in the case of accessibility data in the PROW, prior work demonstrated that many small and underfunded stakeholders confront similar problems in performing relevant, consistent data collection. Having data contributions through collaborative mapping tools and distributed through collaborative data commons like OpenStreetMap (OSM) can improve overall non-motorized transportation assessment and improvement. Travelers stand to benefit through downstream use in routing and wayfinding applications (e.g. Moovel, TransitApp, AccessMap \cite{accessmap, opensidewalks}). The use of collaborative tools and shared data allows for citizen participation in data vetting and collection which can increase trust among agencies and accessibility advocates. Concluding, we believe that an automated sidewalks assessment systems approach is of core importance for the development and management of future urban active transportation networks.

\section*{Acknowledgment}
This work would not have been possible without the vision and leadership of Matthew Weidner at Metro Access, and MV Transportation, Inc. team (with special thanks to Eric A. Peterson, Nik Wilson, and Dwight Sanders). This work was supported by Washington State Department of Transportation Research Grant T1461-47, and Federal Transportation Administration Award No: 693JJ32250014 ITS4US to the Taskar Center for Accessible Technology, and device donations by Nvidia.

\bibliographystyle{IEEEtran}
\bibliography{main}

\end{document}